\documentclass[conference]{IEEEtran}
\IEEEoverridecommandlockouts
\usepackage{cite}
\usepackage{amsmath,amssymb,amsfonts}
\usepackage{algorithmic}
\usepackage{graphicx}
\usepackage{textcomp}
\usepackage{xcolor}
 \usepackage{amstext,epsfig,amssymb,amsbsy,amsmath}
\usepackage{cases}
\newcommand{\mb}{\mathbf}
\usepackage{bm}

\usepackage{amsmath}
\usepackage{booktabs}
\usepackage{cite}
 \usepackage{tikz}
\usetikzlibrary{shapes.geometric}

\pgfdeclareplotmark{mystar}{
    \node[star,star point ratio=2.25,minimum size=5pt,
          inner sep=0pt,draw=blue,solid,fill=blue] {};
}
\usetikzlibrary{plotmarks}
\usetikzlibrary{calc}
\usetikzlibrary{arrows}
\usetikzlibrary{arrows,intersections}
\usetikzlibrary{arrows.meta}
\usetikzlibrary{shapes,decorations}
\usetikzlibrary{positioning}
\usepackage{pgfplots}
\tikzset{sin v source/.style={
  circle,
  draw,
  append after command={
    \pgfextra{
    \draw
      ($(\tikzlastnode.center)!0.5!(\tikzlastnode.west)$)
       arc[start angle=180,end angle=0,radius=0.425ex]
      (\tikzlastnode.center)
       arc[start angle=180,end angle=360,radius=0.425ex]
      ($(\tikzlastnode.center)!0.5!(\tikzlastnode.east)$)
    ;
    }
  },
  scale=1.5,
 }
}
\usetikzlibrary{pgfplots.groupplots, backgrounds, plotmarks, calc, spy, pgfplots.polar, shapes.geometric, arrows, external, patterns}

\tikzset{
    invisible/.style={opacity=0},
    visible on/.style={alt={#1{}{invisible}}},
    alt/.code args={<#1>#2#3}{%
      \alt<#1>{\pgfkeysalso{#2}}{\pgfkeysalso{#3}} 
    },
  }

\tikzset{sin v source/.style={
  circle,
  draw,
  append after command={
    \pgfextra{
    \draw
      ($(\tikzlastnode.center)!0.5!(\tikzlastnode.west)$)
       arc[start angle=180,end angle=0,radius=0.425ex] 
      (\tikzlastnode.center)
       arc[start angle=180,end angle=360,radius=0.425ex]
      ($(\tikzlastnode.center)!0.5!(\tikzlastnode.east)$) 
    ;
    }
  },
  scale=1.5,
 }
}

\pgfplotsset{compat=1.13}

\def\BibTeX{{\rm B\kern-.05em{\sc i\kern-.025em b}\kern-.08em
    T\kern-.1667em\lower.7ex\hbox{E}\kern-.125emX}}
\begin{document}

\title{Learning Optimal Power Flow:\\ Worst-Case Guarantees for Neural Networks
\thanks{The work of A. Venzke was carried out while visiting the Department of Computing and Mathematical Sciences at the California Institute of Technology, Pasadena, CA 91125, USA. The work of A. Venzke and S. Chatzivasileiadis  is  supported  by  the  multiDC  project,  funded  by  Innovation  Fund Denmark, Grant Agreement No. 6154-00020.}
}

\author{\IEEEauthorblockN{Andreas Venzke\IEEEauthorrefmark{1}, Guannan Qu\IEEEauthorrefmark{2}, Steven Low\IEEEauthorrefmark{2} and Spyros Chatzivasileiadis\IEEEauthorrefmark{1}}
\IEEEauthorblockA{\IEEEauthorrefmark{1}Department of Electrical Engineering, Technical University of Denmark (DTU), Kgs. Lyngby, Denmark\\
\IEEEauthorrefmark{2}Department of Computing and Mathematical Sciences, California Institute of Technology, Pasadena, CA 91125, USA \\
E-mail: \{andven, spchatz\}@elektro.dtu.dk, \{gqu, slow\}@caltech.edu}
}

\maketitle

\begin{abstract} 
This paper introduces for the first time a framework to obtain provable worst-case guarantees for neural network performance, using learning for optimal power flow (OPF) problems as a guiding example. Neural networks have the potential to substantially reduce the computing time of OPF solutions. However, the lack of guarantees for their worst-case performance remains a major barrier for their adoption in practice. This work aims to remove this barrier. We formulate mixed-integer linear programs to obtain worst-case guarantees for neural network predictions related to (i)~maximum constraint violations, (ii)~maximum distances between predicted and optimal decision variables, and (iii)~maximum sub-optimality. We demonstrate our methods on a range of PGLib-OPF networks up to 300 buses. We show that the worst-case guarantees can be up to one order of magnitude larger than the empirical lower bounds calculated with conventional methods. More importantly, we show that the worst-case predictions appear at the boundaries of the training input domain, and we demonstrate how we can systematically reduce the worst-case guarantees by training on a larger input domain than the domain they are evaluated on.
\end{abstract}


\section{Introduction}
The optimal power flow (OPF) problem is an essential tool for electricity markets, for power system operation, and planning \cite{cain2012history}. In its standard form, the OPF minimizes an objective function (e.g. generation cost) subject to the power flow equations and the operational constraints (e.g. line limits). As the non-linear AC power flow equations render the AC-OPF problem non-convex \cite{molzahn2019survey}, the linear DC-OPF approximation is often used instead \cite{stott2009dc}. The substantial increase of uncertainty in generation and demand requires to solve OPF repeatedly and closer to real-time, in order to analyze a large number of scenarios; this leads to significant computational challenges \cite{tang2017real}. Neural networks present a promising alternative to conventional optimization solvers, achieving a speed-up of several orders of magnitude \cite{pan2019deepopf,pan2019deepscopf, deka2019learning, chen2020learning,fioretto2019predicting}. However, the lack of any guarantees related to the neural network performance presents a major barrier towards their application in safety-critical systems. In this work, we introduce for the first time a framework to obtain worst-case guarantees for neural networks which predict solutions to DC-OPF problems.

Machine learning including neural networks have been applied to a range of power system applications over the past three decades; for a recent survey please refer to \cite{duchesne2020recent}. The focus of this work is on obtaining guarantees for machine learning approaches such as the ones in \cite{pan2019deepopf,pan2019deepscopf, deka2019learning, chen2020learning,fioretto2019predicting}, which predict solutions to OPF problems and replace the use of conventional optimization solvers. These approaches can result to larger computational speed-ups compared to predicting inactive constraints \cite{pineda2020data} or warm-start points \cite{baker2019learning} that could accelerate conventional optimization solvers. The work in \cite{pan2019deepopf} trains neural networks to directly predict the solution to DC-OPF problems, achieving a speed-up of two orders of magnitude (i.e., 100 times faster).  The same authors extend this framework to include security constraints in \cite{pan2019deepscopf}. The work in \cite{ng2018statistical} proposes an off-line algorithm to identify the sets of active constraints and, based on these, directly computes solutions to DC-OPF problems on-line. The work in \cite{deka2019learning} extends this approach to neural networks predicting the active set. The work in \cite{chen2020learning} demonstrates that both the approaches in \cite{pan2019deepopf} and \cite{ng2018statistical} can fail to predict feasible solutions, i.e., solutions satisfying the power system constraints, and proposes an alternative training procedure to improve the feasibility of the obtained predictions. Using neural networks, the work in \cite{fioretto2019predicting} directly predicts solutions to AC-OPF problems and relies on a penalization of constraint violations during training to improve feasibility.

While the works \cite{pan2019deepopf,pan2019deepscopf, deka2019learning, chen2020learning,fioretto2019predicting} report substantial computational speed-ups and empirically analyse accuracy and feasibility, \emph{no guarantees} for the neural network performance are provided. By evaluating the worst-case performance on the discrete samples for the entire training and test dataset only an empirical lower bound of the worst-case guarantee can be obtained. To the best of our knowledge, this work is the first to introduce a framework that obtains \emph{exact} worst-case guarantees over the \emph{entire} input domain, for neural networks predicting solutions to DC-OPF problems. To this end, we leverage recent advancements in evaluating the adversarial robustness of neural networks using mixed-integer programming \cite{xiao2018training, tjeng2017evaluating, venzke2019verification}. Our previous work \cite{venzke2019verification} focused on power system security assessment and provided performance guarantees for classification neural networks. While these works \cite{xiao2018training, tjeng2017evaluating,venzke2019verification} focus on obtaining \emph{local} robustness certificates that no adversarial examples exist (i.e., input perturbations around a given sample which lead to a wrong classification), in this work we introduce a framework to obtain \emph{global} worst-case guarantees over the entire input domain. The main contributions of our work are:
\begin{enumerate}
\item  We introduce a framework to compute worst-case guarantees for (i)~physical constraint violations, (ii)~maximum distance between predicted and optimal decision variables, and (iii)~sub-optimality, leveraging mixed-integer linear reformulations of neural networks.
\item We demonstrate our methodology on PGLib-OPF networks of up to 300 buses. We show (i) that the worst-case \emph{guarantees over the entire input domain} can be up to an order of magnitude larger than the empirical lower bounds obtained with conventional methods; and (ii) that these worst-case guarantees can be systematically reduced by training on a larger input domain than the domain these neural networks are evaluated on. 
\end{enumerate}

The structure of this paper is as follows: In Section~\ref{Learning_DCOPF}, we formulate the DC-OPF and its Karush–Kuhn–Tucker (KKT) conditions, and explain the neural network architecture and training to predict solutions to DC-OPF problems. In Section~\ref{WC_guarantees}, leveraging mixed-integer reformulations of neural networks, we introduce the framework to compute worst-case guarantees. Section~\ref{Sim_Res} demonstrates our methodology on a range of PGLib-OPF networks. Section~\ref{Conclusion} concludes. The code to reproduce all simulation results is available online \cite{appendix}.

\section{Learning DC-OPF with Neural Networks} \label{Learning_DCOPF}
First, we state the DC-OPF problem and its KKT conditions (which we will use at a later stage in Section \ref{sec:vdist_vopt}), and then we detail the architecture and training process of neural networks predicting solutions to DC-OPF problems.
\subsection{DC Optimal Power Flow (DC-OPF) Formulation}
An electric power grid consists of an $n_{\text{b}}$ number of buses (denoted with the set $\mathcal{N}$) and an $n_{\text{line}}$ number of lines (denoted with the set $\mathcal{L}$). Each line connects a bus $i \in \mathcal{N}$ to another bus $j \in \mathcal{N}$, $(i,j) \in \mathcal{L}$. Set $\mathcal{G}$ (a subset of $\mathcal{N}$) collects the $n_{\text{g}}$ number of buses that have a generator connected to them. The vector $\mb{p_{\text{g}}}$ of size $n_{\text{g}}$ denotes the generator active power output and the matrix $\mb{M}_{\text{g}}$ of size $n_{\text{b}} \times n_{\text{g}}$  maps the generators to the buses. A number $n_{\text{d}}$ of buses has a load connected to them. The vector $\mb{p_{\text{d}}}$ of size $n_{\text{d}}$ denotes the active power demands and the matrix $\mb{M}_{\text{d}}$ of size $n_{\text{b}} \times n_{\text{d}}$  maps the loads to the buses. In the DC-OPF formulation, the voltage magnitudes are assumed to be constant at all buses, and only the voltage angles $\bm{\theta}$ of size $n_{\text{b}}$ are included as variables. The DC-OPF problem can be formulated as:
\begin{alignat}{3}
    \min_{\mb{p_{\text{g}}},\bm{\theta}} \quad & \mb{c}^T \mb{p_{\text{g}}} \label{obj} && \\
    \text{s.t.} \quad & \mb{M}_{\text{g}} \mb{p_{\text{g}}} - \mb{M}_{\text{d}} \mb{p_{\text{d}}} = \mb{B_{\text{bus}}} \bm{\theta}  && \quad : \bm{\lambda} \label{nodal}\\
    & \mb{p_{\text{line}}^{\text{min}}} \leq \mb{B_{\text{line}}} \bm{\theta} \leq \mb{p_{\text{line}}^{\text{max}}}  && \quad : \bm{\mu_{\text{line}}^{\text{min}}}, \bm{\mu_{\text{line}}^{\text{max}}} \label{pline_lim} \\
    & \mb{p_{\text{g}}^{\text{min}}} \leq \mb{p_{\text{g}}} \leq \mb{p_{\text{g}}^{\text{max}}}  && \quad : \bm{\mu_{\text{g}}^{\text{min}}}, \bm{\mu_{\text{g}}^{\text{max}}} \label{pg_lim} 
\end{alignat}
The objective function in \eqref{obj} minimizes the generation cost, with a positive unique linear cost term $\mb{c}$ associated to each generator output. The nodal power balance in \eqref{nodal} ensures that the power generation, power demand and in- and out-going flows are balanced at each bus. The term $\mb{B_{\text{bus}}}$ defines the bus admittance matrix, and the term $\mb{B_{\text{line}}}$ the line admittance matrix. For brevity,  refer to \cite{stott2009dc} for the full details. The active power line flows in \eqref{pline_lim} are a function of the line admittance matrix $\mb{B_{\text{line}}}$ and the voltage angles $\bm{\theta}$. We
fix the voltage angle corresponding to the slack bus $\bm{\theta^{\text{slack}}} = 0$ to remove the trivial non-uniqueness of the obtained DC-OPF solution due to the singularity of the bus admittance matrix $\mb{B_{\text{bus}}}$. The physical constraints comprise minimum and maximum limits on the active line flow in \eqref{pline_lim} and the active power generation in \eqref{pg_lim}, respectively. Each constraint is associated with a dual variable, denoted with $\bm{\lambda}$ for equality constraints and $\bm{\mu}$ for inequality constraints. The KKT conditions of the DC-OPF problem in \eqref{obj}--\eqref{pg_lim} can be written as:
\begin{align}
    \mb{c} - \bm{\mu_{\text{g}}^{\text{min}}} + \bm{\mu_{\text{g}}^{\text{max}}} + \mb{M}_{\text{g}}^T \bm{\lambda} = 0 \label{stat_1} \\
    -\mb{B_{\text{line}}^T} \bm{\mu_{\text{line}}^{\text{min}}}+\mb{B_{\text{line}}^T} \bm{\mu_{\text{line}}^{\text{max}}} -  \mb{B_{\text{bus}}} \bm{\lambda}  = 0 \label{stat_2} \\
    \bm{\mu_{\text{line}}^{\text{min}}} (\mb{p_{\text{line}}^{\text{min}}}-\mb{B_{\text{line}}} \bm{\theta}) = 0,\, \bm{\mu_{\text{line}}^{\text{min}}} (\mb{B_{\text{line}}} \bm{\theta}-\mb{p_{\text{line}}^{\text{max}}}) = 0 \label{compl_slack_1} \\
    \bm{\mu_{\text{g}}^{\text{min}}} (\mb{p_{\text{g}}^{\text{min}}}-\mb{p_{\text{g}}})=0, \, \bm{\mu_{\text{g}}^{\text{max}}} (\mb{p_{\text{g}}}-\mb{p_{\text{g}}^{\text{max}}})=0 \label{compl_slack_2} \\
    \bm{\mu_{\text{g}}^{\text{min}}} \geq 0, \, \bm{\mu_{\text{g}}^{\text{max}}} \geq 0, \bm{\mu_{\text{line}}^{\text{min}}} \geq 0, \, \bm{\mu_{\text{line}}^{\text{max}}} \geq 0 \label{dual_feas} \\
    \eqref{nodal} - \eqref{pg_lim} \label{primal_feas}
\end{align}
The stationarity conditions are described in \eqref{stat_1} and \eqref{stat_2}. The complementary slackness conditions are enforced in \eqref{compl_slack_1} and \eqref{compl_slack_2}. The primal and dual feasibility corresponds to \eqref{primal_feas} and \eqref{dual_feas}, respectively. As the DC-OPF in \eqref{obj}--\eqref{pg_lim} is a linear program, satisfying the KKT conditions is necessary and sufficient for optimality \cite{boyd2004convex}, given the DC-OPF problem is feasible. 

\subsection{Neural Network Architecture and Training}
\begin{figure}
  \def\layersep{1.5cm}
\centering
\resizebox{0.75\columnwidth}{!}{%
\begin{tikzpicture}[shorten >=1pt,->,draw=black!50, node distance=\layersep]
    \tikzstyle{every pin edge}=[<-,shorten <=1pt]
    \tikzstyle{neuron}=[circle,fill=black!25,minimum size=22pt,inner sep=0pt]
    \tikzstyle{input neuron}=[neuron];
    \tikzstyle{output neuron}=[neuron];
    \tikzstyle{hidden neuron}=[neuron];
    \tikzstyle{annot} = [text width=4em, text centered]

    \foreach \name / \y in {1}
        \node[input neuron] (I-\name) at (0,-\y cm - 0.5cm) {$\mb{p_{\text{d}}}$};

    \foreach \name / \y in {1,...,2}
        \path[yshift=0.5cm]
            node[hidden neuron] (H1-\name) at (\layersep,-\y cm) {$\bm{z}^\y_1$};
            
                \foreach \name / \y in {3}
        \path[yshift=0.5cm]
            node[hidden neuron] (H1-\name) at (\layersep,-\y cm) {$\bm{z}^{N_1}_1$};

    \foreach \name / \y in {1,...,2}
        \path[yshift=0.5cm]
            node[hidden neuron,right of=H1-\name] (H2-\name) at (\layersep,-\y cm) {$\bm{z}^\y_2$};
            
                \foreach \name / \y in {3}
        \path[yshift=0.5cm]
            node[hidden neuron,right of=H1-\name] (H2-\name) at (\layersep,-\y cm) {$\bm{z}^{N_2}_2$};
            
    \foreach \name / \y in {1,...,2}
        \path[yshift=0.5cm]
            node[hidden neuron,right of=H2-\name] (H3-\name) at (\layersep+\layersep,-\y cm) {$\bm{z}^\y_3$};
            
                \foreach \name / \y in {3}
        \path[yshift=0.5cm]
            node[hidden neuron,right of=H2-\name] (H3-\name) at (\layersep+\layersep,-\y cm) {$\bm{z}^{N_3}_3$};
            
    \foreach \name / \y in {1,...,2}
        \path[yshift=0.5cm]
            node[hidden neuron,right of=H3-\name] (H4-\name) at (\layersep+\layersep+\layersep,-\y cm) {$\bm{z}^\y_K$};
            
                \foreach \name / \y in {3}
        \path[yshift=0.5cm]
            node[hidden neuron,right of=H3-\name] (H4-\name) at (\layersep+\layersep+\layersep,-\y cm) {$\bm{z}^{N_K}_K$};

    \node[output neuron,right of=H4-2] (O1) {$\mb{\hat{p}_{\text{g}}}$};
    \foreach \source in {1}
        \foreach \dest in {1,...,3}
            \path (I-\source) edge (H1-\dest);
            
                \foreach \source in {1,...,3}
        \foreach \dest in {1,...,3}
            \path (H1-\source) edge (H2-\dest);
            
                \foreach \source in {1,...,3}
        \foreach \dest in {1,...,3}
            \path (H2-\source) edge (H3-\dest);

    \foreach \source in {1,...,3}
        \path (H4-\source) edge (O1);
        
\draw[-,densely dotted, thick, line width = 1pt] (H1-2) -- (H1-3);
\draw[-,densely dotted, thick, line width = 1pt] (H2-2) -- (H2-3);
\draw[-,densely dotted, thick, line width = 1pt] (H3-2) -- (H3-3);
\draw[-,densely dotted, thick, line width = 1pt] (H4-2) -- (H4-3);

\draw[-,densely dotted, thick, line width = 1pt] (H3-1) -- (H4-1);
\draw[-,densely dotted, thick, line width = 1pt] (H3-2) -- (H4-2);
\draw[-,densely dotted, thick, line width = 1pt] (H3-3) -- (H4-3);

    \node[annot,above of=H1-1, node distance=1.2cm] (hl1) {Hidden layer 1};
    \node[annot,above of=H1-1, node distance=0.55cm] (w1b1) {};
    \node[annot,left of=w1b1, node distance=0.85cm] (w1b1t) {$\mb{W}_1,\mb{b}_1$};
    \node[annot,above of=H2-1, node distance=1.2cm] (hl2) {Hidden layer 2};
        \node[annot,above of=H2-1, node distance=0.55cm] (w2b2) {};
    \node[annot,left of=w2b2, node distance=0.75cm] (w2b2t) {$\mb{W}_2,\mb{b}_2$};
    \node[annot,above of=H3-1, node distance=1.2cm] (hl3) {Hidden layer 3};
    \node[annot,above of=H3-1, node distance=0.55cm] (w3b3) {};
        \node[annot,left of=w3b3, node distance=0.75cm] (w3b3t) {$\mb{W}_3,\mb{b}_3$};
    \node[annot,left of=hl1] {Input layer};
    \node[annot,right of=hl3] (hl4)  {Hidden layer K};
        \node[annot,right of=hl4] (hl5)  {Output layer};
        \node[annot,above of=H4-1, node distance=0.55cm] (w4b4) {};
        \node[annot,right of=w3b3, node distance=0.75cm] (w4b4t) {$\mb{W}_K,\mb{b}_K$};
        \node[annot,right of=w4b4, node distance=0.75cm] (w4b4t2) {$\mb{W}_{K+1},\mb{b}_{K+1}$};
\end{tikzpicture}
}
    \caption{Illustration of the neural network architecture to predict the mapping from the active power demand $\mb{p_\text{d}}$ to the optimal generation $\mb{\hat{p}_\text{g}}$: The neural network consists of $K$ hidden layers with $N_k$ neurons each with $k = 1,...,K$. At each neuron of the hidden layers, a ReLU activation function is applied.}
    \label{NN_structure}
\end{figure}
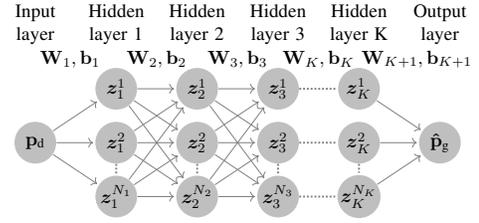
This subsection details the neural network architecture and the training procedure in order to learn the mapping between an instance of the power demand $\mb{p_{\text{d}}}$ and the associated optimal generation dispatch $\mb{p_{\text{g}}}$ of the DC-OPF, see also \eqref{obj}--\eqref{pg_lim}. We assume the power system topology is fixed, i.e., $\mb{B_{\text{bus}}}$ and $\mb{B_{\text{line}}}$ remain constant, and the load domain $\mb{p_{\text{d}}} \in \mathcal{D}$ is restricted to a convex polytope characterized by matrix $\mb{A_\text{d}}$ and vector $\mb{b_\text{d}}$:
\begin{align}
    \mb{A_\text{d}} \mb{p_\text{d}} \leq \mb{b_\text{d}} \label{input_domain}
\end{align}
On this load domain, we assume that the DC-OPF mapping from system demand $\mb{p_{\text{d}}}$ to the optimal generation dispatch $\mb{p_{\text{g}}}$ is unique, i.e., a singleton. It is shown in \cite{zhou2019optimal} that the DC-OPF solution is unique almost surely in an appropriate space. This is a requirement for the learning task as the neural network predicts one optimal generation dispatch $\mb{\hat{p}_{\text{g}}}$ for the active power demand input.
The neural network architecture to learn this mapping is illustrated in Fig.~\ref{NN_structure}. The architecture comprises a number $K$ of fully-connected hidden layers, each of which consists of a number of neurons $N_k$ with $k = 1,...,K$. The neural network input vector is the active power demand $\mb{p_{\text{d}}}$ and the output vector is the prediction of the optimal generation dispatch $\mb{\hat{p}_{\text{g}}}$. Note that the entry of $\mb{\hat{p}_{\text{g}}}$ corresponding to the slack bus $(\mb{\hat{p}_\text{g}})^{\text{slack}}$ is not predicted by the neural network as it is not an independent variable. The slack bus generation is defined by the difference in predicted generation and demand:
\begin{align}
    (\mb{\hat{p}_\text{g}})^{\text{slack}} = \sum_{i \in \mathcal{N}} (\mb{M_\text{d}} \mb{p_\text{d}})^i - \sum_{i \in \mathcal{G} \setminus \text{slack}} (\mb{\hat{p}_\text{g}})^i \label{slack_bal}
\end{align} 
The superscripts are used to denote the corresponding entries of the vectors. The input to the first and subsequent hidden layers $\mb{\hat{z}}_k$ of the neural network is defined as:
\begin{alignat}{3}
 \mb{\hat{z}}_{1} = & \mb{W}_{1} \mb{p_\text{d}} + \mb{b}_{1} && \label{NN_1a}\\ 
    \mb{\hat{z}}_{k+1} = & \mb{W}_{k+1} \mb{z}_k + \mb{b}_{k+1} && \quad \forall k  = 1, ..., K-1 \label{NN_1b}
\end{alignat}
The weight matrices $\mb{W}_{k}$ have dimensions $N_{k+1} \times N_k$ and the bias vector $\mb{b}$ has dimension $N_{k+1}$. 
Each neuron in the hidden layer applies a non-linear activation function to the input. In the following, we use the ReLU activation function, which is used by the majority of neural network applications in recent years, as it has been found to accelerate neural network training  \cite{glorot2011deep}: 
\begin{align}
    \mb{z}_k^i = \max (\mb{\hat{z}}_{k}^i,0) & \quad \forall k  = 1, ..., K \quad  \forall i = 1, ...,N_k \label{NN_2}
\end{align}
The ReLU activation function in \eqref{NN_2} outputs $0$ if the input is negative, otherwise it propagates the input. Note that the $\max$ operator is applied element-wise to the vector $\mb{\hat{z}}_k$. The predicted generator dispatch of the neural network can be evaluated as follows:
\begin{align}
    (\mb{\hat{p}_\text{g}})^{\text{nsg}} = \mb{W}_{K+1} \mb{z}_{K} + \mb{b}_{K+1} \label{NN_3}
\end{align}
The term $ (\mb{\hat{p}_\text{g}})^{\text{nsg}}$ denotes the $n_{\text{g}}-1$ entries of $\mb{\hat{p}_{\text{g}}}$ that do not correspond to the slack bus. 
To train neural networks, the first step is to create a dataset of demand instances $\mb{p_\text{d}} \in \mathcal{D}$ and their corresponding optimal generation $\mb{p_\text{g}}$ by e.g. using historical data and simulation tools. The obtained dataset is split into a training and test set. Then, during neural network training, the weight matrices $\mb{W}$ and biases $\mb{b}$ are optimized using stochastic gradient descent to minimize a loss function, e.g. the mean squared error between the prediction $\mb{\hat{p}_\text{g}}$ and the training dataset $\mb{p_\text{g}}$. In previous works (e.g. \cite{pan2019deepopf, chen2020learning}), the performance of the trained neural network is evaluated on the test set using \emph{statistical} metrics such as accuracy or share of feasible instances. This procedure does not provide any \emph{guarantees} related to the worst-case performance of the trained neural network over the entire input domain $\mb{p_\text{d}} \in \mathcal{D}$.

\section{Worst-Case Guarantees for Neural Networks} \label{WC_guarantees}
We first state the mixed-integer reformulation of trained neural networks and address issues related to scalability. Then, we introduce our framework to compute  worst-case guarantees.
\subsection{Mixed-Integer Reformulation of Trained Neural Networks}  \label{MILP_TNN}
To include the trained neural network equations in an optimization framework, we follow the work in  \cite{tjeng2017evaluating} and reformulate the maximum operator in the ReLU activations \eqref{NN_2} using binary variables $\mb{b}_k \in \{0,1\}^{N_k}$ for all $k = 1, ...,K$ and suitable minimum and maximum bounds on the neuron output $\hat{\mb{z}}^{\text{min}}$ and $\hat{\mb{z}}^{\text{max}}$:
\begin{alignat}{3}
\mb{z}_k^i  &\leq \hat{\mb{z}}_k^i - \hat{\mb{z}}^{\text{min},i}_k  (1-\mb{b}_k^i) && \, \, \forall k = 1, ...,K \,  \forall i = 1, ...,N_k \label{ReLU1} \\ 
\mb{z}_k^i  &\geq \hat{\mb{z}}_k^i && \, \, \forall k = 1, ...,K \,  \forall i = 1, ...,N_k  \label{ReLU2}   \\
\mb{z}_k^i  &\leq \hat{\mb{z}}^{\text{max},i}_k \mb{b}_k^i && \,  \,\forall k = 1, ...,K  \,  \forall i = 1, ...,N_k  \label{ReLU3}  \\
\mb{z}_k^i   &\geq \mb{0}  && \, \forall k = 1, ...,K \, \,  \forall i = 1, ...,N_k  \label{ReLU4}  \\
\mb{b}_k &\in \{0,1\}^{N_k}  && \, \, \forall k = 1, ...,K\label{ReLUe}
\end{alignat}
Observe that $\mb{\hat{z}}_k^i$ refers to the neuron (ReLU) input and $\mb{z}_k^i$ to the neuron (ReLU) output. Note that the minimum and maximum bounds on the neuron output $\hat{\mb{z}}^{\text{min}}$ and $\hat{\mb{z}}^{\text{max}}$ have to be chosen large enough to not be binding and as small as possible to facilitate tight bounds for the branch-and-bound algorithm. In case the input to the $i$-th neuron in layer $k$ is $\mb{\hat{z}}_k^i \leq 0$ then the corresponding binary variable $\mb{b}_k^i$ is 0 and \eqref{ReLU3} and \eqref{ReLU4} constrain the neuron output $\mb{z}_k^i$ to 0. The constraints in \eqref{ReLU1} and \eqref{ReLU2} are non-binding if $\mb{\hat{z}}_k^i  < 0$ holds. If the input to the neuron is $\mb{\hat{z}}_k^i \geq 0$, then the binary variable is 1 and \eqref{ReLU1} and \eqref{ReLU2} constrain the neuron output $\mb{z}_k^i$ to the input $\mb{\hat{z}}_k^i$. The constraints in \eqref{ReLU3} and \eqref{ReLU4} are non-binding if $\mb{\hat{z}}_k^i > 0$ holds. 
 
 As this reformulation introduces one binary variable for each neuron in the hidden layers, we use a combination of the works in \cite{tjeng2017evaluating} and \cite{xiao2018training} and employ three techniques to maintain scalability of the resulting mixed-integer linear programs (MILPs). First, we sparsify the weight matrices $\mb{W}$ during training, i.e., we gradually enforce a defined share of entries to be zero. Second, we apply the concept of ReLU stability \cite{xiao2018training}: All neurons for which the activation is always active or always inactive on both the training and test set are fixed to this status in the MILP reformulation, and the corresponding binaries are eliminated. Third, we use several techniques to compute increasingly tighter bounds $\hat{\mb{z}}^{\text{min}}$ and $\hat{\mb{z}}^{\text{max}}$. We initialize the bounds using interval arithmetic (for details see \cite{tjeng2017evaluating}). Then, to compute tighter bounds, we minimize and maximize the output of each neuron $\mb{z}_k^i$ subject to the linear relaxation of the trained neural network \eqref{NN_1a}, \eqref{NN_1b}, \eqref{NN_3}--\eqref{ReLUe}, and subject to the restricted input domain in \eqref{input_domain}. Note that for the linear relaxation only we relax the binary variables $\mb{b}_k$ to continuous variables between $\mb{0}$ and $\mb{1}$. Finally, we repeat this step using the full MILP formulation of the trained neural networks. 
 As a result, we obtain tightened bounds $\hat{\mb{z}}^{\text{min}}$ and $\hat{\mb{z}}^{\text{max}}$ for the branch-and-bound algorithm. Note that in the following, when solving MILPs to obtain worst-case guarantees, we always solve the full MILP formulation and do not use a relaxation.
 
\subsection{Worst-Case Guarantees for Constraint Violations}
The mixed-integer reformulation of trained neural networks allows us to formulate optimization problems to obtain worst-case guarantees for the physical constraint violation. We define the maximum violation of the constraints on active generator power $\nu_{\text{g}}$ in \eqref{pg_lim} and on active line flows $\nu_{\text{line}}$ in \eqref{pline_lim} as:
\begin{align}
\nu_{\text{g}} & = \max(\mb{\hat{p}_\text{g}}-\mb{p_\text{g}^{\text{max}}},\mb{p_\text{g}^{\text{min}}}-\mb{\hat{p}_\text{g}},\mb{0}) \label{vg} \\
\nu_{\text{line}}  & = \max(|\mb{B_{\text{line}}} \mb{\tilde{B}_{\text{bus}}^{-1}} (\mb{M}_{\text{g}} \mb{\hat{p}_\text{g}}- \mb{M}_{\text{d}}\mb{p_\text{d}})^{\text{nsb}}| -  \mb{p_{\text{line}}^{\text{max}}},\mb{0}) \label{vline} 
\end{align}
 The term `$\text{nsb}$' denotes all buses except the slack bus. To compute the maximum constraint violation of the line flow in \eqref{vline}, we compute the line flow based on the neural network prediction $\mb{\hat{p}_\text{g}}$ and system loading $\mb{p_{\text{d}}}$. 
 To this end, we remove the column and row from the bus admittance matrix and invert the resulting reduced bus admittance matrix $\mb{\tilde{B}_{\text{bus}}}$, inserting \eqref{nodal} in \eqref{pline_lim}. Note that the product $\mb{B_{\text{line}}} \mb{\tilde{B}_{\text{bus}}^{-1}}$ is the well-known ``Power Transfer Distribution Factors'' (PTDF) matrix; please refer to \cite{LectureNotes_OPF_SCH} for more details.
 In both \eqref{vg} and \eqref{vline}, we take the overall non-negative maximum over the violations. Note that we take the absolute value $|\cdot|$ of the line flow in \eqref{vline}. In previous works, these metrics have only been evaluated empirically on the datasets. Here, to compute the worst-case generator constraint violation for the entire defined input domain, we solve:
\begin{align}
\max_{\mb{\hat{p}_\text{g}},\mb{p_\text{d}},\mb{b},\mb{z},\mb{\hat{z}},\nu_{\text{g}}} \, \, &  \nu_{\text{g}} \label{wc_constr_viol_obj} \\
\text{s.t.} \, \, & \eqref{input_domain}-\eqref{NN_1b}, \eqref{NN_3}, \eqref{ReLU1}-\eqref{ReLUe}, \eqref{vg} \label{wc_constr_viol} 
\end{align}
Similarly, to compute the maximum line constraint violation $\nu_{\text{line}}$, we maximize $\nu_{\text{line}}$ subject to \eqref{wc_constr_viol}, replacing \eqref{vg} with \eqref{vline}. As the input domain in \eqref{input_domain} is a convex polytope and we reformulate the max-operators in \eqref{vg} and \eqref{vline} using integer variables, the optimization problem \eqref{wc_constr_viol_obj}--\eqref{wc_constr_viol} can be cast as MILP. If the MILP is solved to zero MILP gap, i.e., to global optimality, then the bound is exact, and we obtain the provable guarantee that no input $\mb{p_{\text{d}}} \in \mathcal{D}$ to the neural network exist which will lead to constraint violations larger than the obtained values of $\nu_{\text{g}}$ and  $\nu_{\text{line}}$. At the same time, the obtained values of $\mb{p_{\text{d}}}$ are the neural network inputs which lead to the maximum constraint violations. If the MILP is solved to a non-zero optimality gap, then we obtain an upper bound on the worst-case violations $\nu_{\text{g}}$ and $\nu_{\text{line}}$. If, additionally, the MILP solver identifies a feasible solution, then the values of $\nu_{\text{g}}$ and  $\nu_{\text{line}}$ corresponding to the feasible solution serve as a lower bound on the worst-case violations. Note that in the simulation results in Section~\ref{Sim_Res}, we solve all MILPs to zero optimality gap. 
\subsection{Worst-Case Guarantees for Distance of Predicted to Optimal Decision Variables and for Sub-Optimality}
\label{sec:vdist_vopt}
In the following, we formulate optimization problems to obtain (i)~worst-case guarantees for the maximum distance between the predicted and the optimal decision variables $\nu_{\text{dist}}$ and (ii)~worst-case guarantees for the sub-optimality of the cost function $\nu_{\text{opt}}$ resulting from the predicted solution: 
\begin{align}
    \nu_{\text{dist}} & = \max(|\tfrac{|\mb{\hat{p}_{\text{g}}}-\mb{p_{\text{g}}}|}{\mb{p_{\text{g}}^{\text{max}}}-\mb{p_{\text{g}}^{\text{min}}}}|) \label{vdist}  \\
    \nu_{\text{opt}} & =  \mb{c}^T(\mb{\hat{p}_{\text{g}}}-\mb{p_{\text{g}}}) \label{vopt} 
\end{align}
The term $\mb{p_{\text{g}}}$ denotes the optimal solution to the DC-OPF problem for a given input loading $\mb{p_{\text{d}}}$. We normalize the distance $\nu_{\text{dist}}$ element-wise by the corresponding generator limits and compute the maximum over all generator set-points. The distance $\nu_{\text{dist}}$ characterizes for the entire input domain the largest mismatch of all generator set-points between the prediction of the neural network and the ground-truth DC-OPF solution. We formulate the following bi-level problem to compute the worst-case distance $\nu_{\text{dist}}$:
\begin{align}
\max_{\mb{\hat{p}_{\text{g}}},\mb{p_{\text{g}}},\mb{p_{\text{d}}},\mb{b},\mb{z},\mb{\hat{z}},\nu_{\text{dist}} } \, \, &  \nu_{\text{dist}} \label{bilevelstart} \\
\text{s.t.} \, \, & \eqref{input_domain}-\eqref{NN_1b}, \eqref{NN_3}, \eqref{ReLU1}-\eqref{ReLUe} , \eqref{vdist} \label{constr_wc_dist_1} \\
& \mb{p_{\text{g}}} \in \arg\min_{\bm{p_{\text{g}}},\bm{\theta}} \{\eqref{obj} \text{ s.t. } \eqref{nodal} - \eqref{pg_lim} \} \label{bilevelend} 
\end{align}
The lower-level comprises the DC-OPF formulation and defines the optimal generation $\mb{p_{\text{g}}}$ as a function of the load input $\mb{p_{\text{d}}}$. The upper-level problem maximizes the distance of the predicted to the optimal solution of the DC-OPF for the defined load input domain. We replace the lower-level problem with its KKT conditions and rewrite the optimization problem:
\begin{align}
\max_{\mb{\hat{p}_{\text{g}}},\mb{p_{\text{g}}},\mb{p_{\text{d}}},\mb{b},\mb{z},\mb{\hat{z}},\nu_{\text{dist}} ,\bm{\theta}, \bm{\lambda},\bm{\mu}} \, \, &  \nu_{\text{dist}} \label{wc_dist_obj}\\
\text{s.t.} \, \, & \eqref{constr_wc_dist_1}, \eqref{stat_1}-\eqref{primal_feas} \label{wc_dist_constr} 
\end{align}
By maximizing $\nu_{\text{opt}}$ in the the objective function and replacing \eqref{vdist} with \eqref{vopt} we can compute worst-case guarantees for the sub-optimality of the predicted solution. To achieve tractability of this formulation, we reformulate the non-linear complementary slackness conditions \eqref{compl_slack_1} and \eqref{compl_slack_2} in \eqref{wc_dist_constr} using the Fortuny-Amat McCarl linearization \cite{fortuny1981representation}:
\begin{alignat}{4}
\mb{p_{\text{line}}^{\text{min}}}-\mb{B_{\text{line}}} \bm{\theta} & \geq - \mb{r_{\text{line}}^{\text{min}}} \mb{M_{\text{line}}^{\text{min}}}, && \, \, \bm{\mu_{\text{line}}^{\text{min}}} \leq && (1 - \mb{r_{\text{line}}^{\text{min}}} ) \mb{M_{\text{line}}^{\text{min}}} \label{FAM_1} \\
\mb{B_{\text{line}}} \bm{\theta}-\mb{p_{\text{line}}^{\text{max}}} & \geq - \mb{r_{\text{line}}^{\text{max}}} \mb{M_{\text{line}}^{\text{max}}}, && \, \, \bm{\mu_{\text{line}}^{\text{max}}} \leq && (1 - \mb{r_{\text{line}}^{\text{max}}} ) \mb{M_{\text{line}}^{\text{max}}}  \\
\mb{p_{\text{g}}^{\text{min}}}-\mb{p_{\text{g}}} & \geq - \mb{r_{\text{g}}^{\text{min}}} \mb{M_{\text{g}}^{\text{min}}}, && \,\, \bm{\mu_{\text{g}}^{\text{min}}}  \leq && (1 - \mb{r_{\text{g}}^{\text{min}}} ) \mb{M_{\text{g}}^{\text{min}}} \\
\mb{p_{\text{g}}}-\mb{p_{\text{g}}^{\text{max}}} & \geq - \mb{r_{\text{g}}^{\text{max}}} \mb{M_{\text{g}}^{\text{max}}}, && \, \, \bm{\mu_{\text{g}}^{\text{max}}}  \leq && (1 - \mb{r_{\text{g}}^{\text{max}}} ) \mb{M_{\text{g}}^{\text{max}}} \label{FAM_2}
\end{alignat}
This models the complementary slackness conditions with one binary variable $\mb{r}$ and one large non-binding constant $\bm{M}$ for each condition. Note that the constant $\bm{M}$ has to be chosen sufficiently large to not be binding, while at the same time small enough to maintain numerical well-conditioning of the mixed-integer program. For details on bi-level programming and this reformulation, please refer to \cite{dempe2015bilevel}. The resulting optimization problem is a MILP which includes integer variables related to the reformulation of the neural network, and related to the reformulation of the lower-level problem. If this MILP is solved to zero MILP gap (and if constraint qualifications for global optimality to the bi-level problem are satisfied \cite{dempe2015bilevel}), the bound is exact and we obtain the provable guarantee that no input $\mb{p_{\text{d}}} \in \mathcal{D}$ exist with distances or sub-optimality larger than the obtained values of $\nu_{\text{dist}}$ and $\nu_{\text{opt}}$.
\section{Simulation \& Results} \label{Sim_Res}
\begin{table}[]
    \caption{Test Case Characteristics}
    \centering
    \begin{tabular}{c c c c c c}
    \toprule
 Test cases   & $n_{\text{d}}$ & $n_{\text{g}}$ & $n_{\text{b}}$ & $n_{\text{line}}$  & Max. loading  \\
\midrule
\textit{case9}        & 3 & 3& 9& 9 & 315.0~MW \\
\midrule
 \textit{case30}     & 21 & 2& 30 & 41& 283.4~MW  \\
 \midrule
 \textit{case39}   & 21  & 10& 39 & 46 & 6254.2~MW  \\
 \midrule
 \textit{case57}     &  42& 4 & 57 & 80 & 1250.8~MW  \\
 \midrule
 \textit{case118}   &  99 & 19 & 118 & 186 & 4242.0~MW \\
 \midrule
 \textit{case162}   & 113 & 12 & 162 & 284  & 7239.1~MW \\ 
  \midrule
 \textit{case300}   & 199 & 57 & 300 & 411 & 23525.9~MW \\
         \bottomrule
    \end{tabular}
    \label{Table_Characteristics}
\end{table}
We demonstrate our methodology on a range of PGLib-OPF networks v19.05 of up to 300 buses from \cite{babaeinejadsarookolaee2019power}. The test case characteristics are listed in Table~\ref{Table_Characteristics}. The \textit{case9} is taken from MATPOWER \cite{zimmerman2010matpower}. We assume that the input domain~$\bm{p_\text{d}} \in \mathcal{D}$ in \eqref{input_domain} is defined as $0.6\,\bm{p_{\text{d}}^{\text{max}}} \leq \bm{p_{\text{d}}} \leq 1.0\,\bm{p_{\text{d}}^{\text{max}}}$, i.e. each load can fluctuate individually from 60\% to 100\% of its maximum loading. Note that the maximum loading level $\bm{p_{\text{d}}^{\text{max}}}$ is defined according to \cite{babaeinejadsarookolaee2019power,zimmerman2010matpower}, and the sum of the maximum loading is shown in Table~\ref{Table_Characteristics}. We did not consider loading levels larger than 100\% as we observed that this frequently leads to infeasibility of the DC-OPF problem. This would require load shedding and represents an abnormal system situation. To create the datasets, we use Latin hypercube sampling \cite{mckay1979comparison}, draw 100'000 samples from the input domain $\mathcal{D}$, and solve a DC-OPF for each of the samples using MATPOWER\cite{zimmerman2010matpower} to generate the corresponding optimal solutions. Out of these input-output pairs we use 80\% for training and 20\% for testing. 

The neural network architecture comprises three hidden layers with 50 neurons each. As we will demonstrate (and has also been shown in \cite{chen2020learning}), the size of this architecture is sufficient to obtain low generalization errors of the neural networks on the unseen test set. As described in Section~\ref{MILP_TNN}, we sparsify the neural network during training by gradually setting the smallest weight entries to zero until 80\% of the weight entries are zero; that means that only 20\% of weight entries are non-zero at the end of training. We use TensorFlow \cite{tensorflow2015-whitepaper} for neural network training with the following specifications. During training, we minimize the mean squared error between the neural network prediction and the true optimal solutions. We define the maximum number of training epochs to 250 and split the dataset into 2000 batches. We use early stopping and we recover the neural network weights and biases that achieved the lowest generalization error on the test set. As the neural network training is highly non-linear, we repeat the training and evaluation process 5 times, and report averaged values for all simulation results. We formulate the MILPs in YALMIP \cite{lofberg2004yalmip} and solve them using Gurobi. For the Fortuny-Amat McCarl linearization in \eqref{FAM_1}--\eqref{FAM_2} we choose all constants $\bm{M}$ to be $10^5$. After solving the MILPs, we verify that the complementary slackness conditions are satisfied and the constants are non-binding. All computational experiments are carried out on a laptop with i7-7820HQ~CPU~@2.90~GHz, 32~GB RAM and GeForce~940MX GPU. The code to reproduce all simulation results is available online \cite{appendix}.
\subsection{Neural Network Performance}
In the following, we evaluate the performance of the trained neural networks with four metrics: The maximum generator and line constraint violations $\nu_{\text{g}}$, $\nu_{\text{line}}$ defined in \eqref{vg} and \eqref{vline}, the distance of the predicted to the optimal decision variables $\nu_{\text{dist}}$ defined in \eqref{vdist},and the sub-optimality $\nu_{\text{opt}}$  defined in \eqref{vopt}. Note that we normalize the sub-optimality with respect to the generation cost of the 100\% loaded system state. 
\subsubsection{Performance Averaged over Test Set Samples}
\begin{table}[]
    \caption{Performance Averaged over Test Set Samples}
    \centering
    \begin{tabular}{c c c c c c }
    \toprule
     Test cases   & MAE &  $\nu_{\text{g}}$ & $\nu_{\text{line}}$ & $\nu_{\text{dist}}$ & $\nu_{\text{opt}}$ \\
     &  (\%) & (MW) & (MW) & (\%) & (\%) \\
\midrule
\textit{case9}     & 0.04  & 0.07 & 0.02 & 0.06 & 0.04 \\
\midrule
 \textit{case30}     & 0.03 &  0.00 & 0.01 & 0.03 & -0.00 \\
 \midrule
 \textit{case39} &  0.07 & 0.71 & 1.02 & 0.30 & 0.00 \\
 \midrule
 \textit{case57}    & 0.01 & 0.24 & 0.00 & 0.03 & -0.01 \\
 \midrule
 \textit{case118}   & 0.31 &  8.21 & 1.35 & 3.35 & 0.00 \\
 \midrule
 \textit{case162}   & 0.61 & 9.11 & 2.07 & 4.08 & 0.01 \\
  \midrule
 \textit{case300}  & 0.90 & 15.33 & 96.13 & 18.01 & -0.02 \\
         \bottomrule
    \end{tabular}
    \label{Table_Test_Set_Performance}
\end{table}
In Table~\ref{Table_Test_Set_Performance}, we show the performance of the trained neural networks averaged over the unseen test dataset samples. The mean absolute error (MAE) of the predicted generation dispatch evaluates to less than 1\% (normalized by the generator limits as in \eqref{vdist}), indicating satisfactory generalization capability of the neural networks. The averaged largest violation of active generator and line limits are less than 0.5\% of the total maximum system loading in Table~\ref{Table_Characteristics}. The averaged largest distances of the predicted and optimal generator dispatch $\nu_{\text{dist}}$ are less than 1\% for the first four test cases, and increases up to 18\% for \textit{case300}. Note that the latter corresponds to the maximum over the vector $\mb{\hat{p}_g}$ of 57 predicted generator set-points. The averaged sub-optimality $\nu_{\text{opt}}$ of the predicted solutions is negligible. Note that the sub-optimality measure can be negative if constraints are violated. The average performance on the test set indicates satisfactory neural network performance. In the following, however, we demonstrate that the worst-case guarantees for these four metrics can be up to two orders of magnitude larger than the average performance on the test set (reported in Table~\ref{Table_Characteristics}). 
\subsubsection{Worst-Case Guarantees for Constraint Violations}
\begin{table}[]
    \caption{Worst-Case Guarantees for Physical Constraint Violations}
    \centering
    \begin{tabular}{c | c c | c c c c}
    \toprule
    & \multicolumn{2}{c}{Emp. lower bound} & \multicolumn{4}{c}{Worst-case guarantee} \\
    \midrule
 Test cases   & $\nu_{\text{g}}$ & $\nu_{\text{line}}$ & \multicolumn{2}{c}{$\nu_{\text{g}}$} & \multicolumn{2}{c}{$\nu_{\text{line}}$} \\
 & (MW) & (MW) & (MW) & (ratio) & (MW) & (ratio)  \\
\midrule
 \textit{case9}     & 2.5 & 1.8 & 2.8 & 1.1x & 1.9 & 1.1x\\
\midrule
\textit{case30}     & 1.7 & 0.6 & 3.6 & 2.1x & 3.1 & 4.9x \\
 \midrule
\textit{case39}     & 51.9 & 37.2& 270.6 & 5.2x & 120.0& 3.2x\\
  \midrule
\textit{case57}     & 4.2  & 0.0 & 23.7 & 5.6x & 0.0 & -- \\
  \midrule
\textit{case118}     & 149.4 & 15.6 & 997.8 & 6.7x & 510.8 & 32.7x\\
 \midrule
\textit{case162}     & 228.0 & 180.0 & 1563.3 & 6.9x & 974.1 & 5.4x\\
  \midrule
\textit{case300}     & 474.5 & 692.7 & 3658.5 & 7.7x & 3449.3  & 5.0x \\
         \bottomrule
    \end{tabular}
    \label{Table_WC_Constr_Viol}
\end{table}
We first compute the worst-case constraint violations on the entire data set, i.e. on all training and test set samples. This serves as an empirical lower bound on the worst-case guarantees. Then, using the mixed-integer linear reformulation of the trained neural networks, we solve the MILPs in \eqref{wc_constr_viol_obj}--\eqref{wc_constr_viol} to compute the corresponding worst-case guarantees.  
In Table~\ref{Table_WC_Constr_Viol}, we compare the obtained empirical lower bounds with the worst-case guarantees related to the violation of the generator constraints $\nu_{\text{g}}$ and of the transmission line constraints $\nu_{\text{line}}$. First, we find that the worst-case guarantees for constraint violations can be substantial.  Table~\ref{Table_WC_Constr_Viol} shows the violations in MW-values. In percentage, the violations are on average 8.1\% and up to 23.5\% (\textit{case118}) of the maximum system loading shown in Table~\ref{Table_Characteristics} for each case. Second, the worst-case guarantees are on average 6.7 times and up to 32.7 times larger than the empirical lower bounds (the empirical lower bounds are obtained by evaluating the worst-case performance on the discrete samples of the entire training and test dataset; if we only consider the test set, then the worst-case guarantees are on average 255.2 times larger than the performance shown in Table~\ref{Table_Test_Set_Performance}). For the \textit{case57} system, we obtained a certificate that no input inside the input domain exists which can lead to a violation of the line constraints. Overall, these findings highlight that by only considering the performance on the dataset, the worst-case performance can be significantly underestimated, posing a risk for real-time deployment if we do not take appropriate mitigation measures. At the same time, our framework allows to obtain a provable exact certificate on the worst-case performance of neural networks. 

By analyzing the solutions, we identified that for 18 out of the 35 evaluations (5 neural networks trained for each test case), the worst-case generator violation ($\nu_\text{g}$) occurs for the slack bus generator, as this generator has to compensate for the mismatch in predicted generation and load. For the line limits, the worst-case violations occurred on a line directly connected to the slack bus for 24 out of the 35 evaluations. Averaged over the 7 test cases and 5 runs for each test case, it takes 3.4 minutes to compute the tightened bounds for the mixed-integer reformulation in \eqref{ReLU1} and \eqref{ReLU3}, and 1.4 minutes to solve both the MILP to zero MILP gap and compute the worst-case guarantees. Based on the activation patterns on the entire dataset, on average, 17.1\% of the ReLU activations are fixed to be active and 39.4\% are fixed to be inactive for solving the MILPs (as described in Section~\ref{MILP_TNN} about ReLU stability).

\subsubsection{Worst-Case Guarantees for Distance of Predicted to Optimal Decision Variables and for Sub-Optimality}
\begin{table}[]
    \caption{Worst-Case Guarantees for (i) Distance of Predicted to Optimal Decision Variables and (ii) Sub-Optimality}
    \centering
    \begin{tabular}{c | c c | c c c c}
    \toprule
    & \multicolumn{2}{c}{Emp. lower bound} & \multicolumn{4}{c}{Worst-case guarantee} \\
    \midrule
 Test cases   & $\nu_{\text{dist}}$ & $\nu_{\text{opt}}$ & \multicolumn{2}{c}{$\nu_{\text{dist}}$} & \multicolumn{2}{c}{$\nu_{\text{opt}}$ } \\
& (\%) & (\%) & (\%) & (ratio) & (\%) & (ratio)  \\
\midrule
 \textit{case9}     & 1.2 & 3.3 & 1.4& 1.2x& 3.8 & 1.1x\\
\midrule
 \textit{case30}     & 2.0 & 0.6 & 6.4& 3.2x & 2.5 & 3.8x \\
 \midrule
 \textit{case39}     & 6.2 & 0.6 & 64.4 & 10.4x & 3.1  &  4.9x \\
 \midrule
  \textit{case57}     & 0.5 & 0.2 & 18.6 & 37.9x & 1.8 & 8.1x \\
 \midrule
  \textit{case118}     & 35.0 & 0.2 &  265.7 & 7.6x & 1.6 &6.5x\\
         \bottomrule
    \end{tabular}
    \vspace{-0.2cm}
    \label{Table_WC_dist}
\end{table}
In the next step, in Table~\ref{Table_WC_dist}, for the same trained neural networks and using the same procedure as in Table~\ref{Table_WC_Constr_Viol}, we compare the obtained empirical lower bounds and worst-case guarantees related to (i)~the maximum distance between the predicted and the optimal decision variables $\nu_{\text{dist}}$ and (ii)~the sub-optimality $\nu_{\text{opt}}$. For these two metrics, we also observe that the worst-case guarantees can be substantial; they are on average 8.5 times and up to 37.9 times larger than the empirical lower bounds which are obtained by calculating the worst-case neural network performance on the discrete dataset samples. 
By analyzing the solutions for the metric $\nu_{\text{dist}}$, we identified that for 12 out of the 25 evaluations, the worst-case distance between the neural network prediction and the optimal solution occurs for the slack bus generator. For the first four test cases, on average, it takes 0.3 minutes to solve both the MILPs to zero MLIP gap. For the \textit{case118}, the average computational time increases to 25.6 minutes to solve both the MILPs to compute $\nu_{\text{dist}}$  and $\nu_{\text{opt}}$ to zero MILP gap. Note that the computational complexity increases as the KKT conditions of the DC-OPF problem are included in \eqref{wc_dist_obj} -- \eqref{wc_dist_constr}. For the \textit{case162} and \textit{case300}, the MILPs could not be solved to a zero MILP gap  within 3 hours. Improving the tractability using decomposition techniques and validating the satisfaction of constraint qualifications for global optimality to the bi-level program in \eqref{bilevelstart} -- \eqref{bilevelend} are subject of our future work \cite{dempe2015bilevel}. 
\subsubsection{Input Domain Reduction}
\begin{figure}
    \begin{footnotesize}
%
%
\definecolor{mycolor1}{rgb}{1.00000,1.00000,0.00000}%
\begin{tikzpicture}

\begin{axis}[%
width=8.5cm,
height=2.8cm,
xmin=0,
xmax=0.5,
xlabel style={font=\color{white!15!black}},
xlabel={(a) \textit{case39}: $\text{Input domain reduction }\delta$ (--) },
ymin=0,
ymax=100,
ylabel style={font=\color{white!15!black}},
ylabel={Guarantee $\nu$ (\%)},
  xtick={0, 0.1,0.2,0.3,0.4,0.5}, 
  xticklabels={0, 0.04,  0.08, 0.12, 0.16,   0.20},
axis background/.style={fill=white},
legend pos=north east,
    legend style={draw=white!15!black,legend cell align=left},
    legend columns=4
]
\addplot [color=red, mark=x, mark options={solid, red},line width =1pt]
  table[row sep=crcr]{%
0	100\\
0.01	95.7249603711599\\
0.05	78.6677149345259\\
0.1	57.6157646877889\\
0.2	19.5923140496445\\
0.3	6.22391037558576\\
0.4	3.00010049803359\\
0.5	0.888981863439581\\
};
\addlegendentry{$\nu_{\text{line}}$}

\addplot [color=blue, mark=x, mark options={solid, blue},line width =1pt]
  table[row sep=crcr]{%
0	100\\
0.01	94.8621237535875\\
0.05	75.5862094323922\\
0.1	49.9819664767147\\
0.2	2.22949802249841\\
0.3	1.37816056589014\\
0.4	0.833377656840963\\
0.5	0.0589226092399364\\
};
\addlegendentry{$\nu_{\text{g}}$};

\addplot [color=green, mark=x, mark options={solid, green},line width =1pt]
  table[row sep=crcr]{%
0	100\\
0.01	94.7923665100223\\
0.05	73.4059108142645\\
0.1	46.7971074322319\\
0.2	5.89396529284167\\
0.3	1.87734435741503\\
0.4	0.954211755803644\\
0.5	0.331692052360007\\
};
\addlegendentry{$\nu_{\text{dist}}$};

      \addplot [color=black, mark=x, mark options={solid, black},line width =1pt]
  table[row sep=crcr]{%
0	100\\
0.01	93.0598344204222\\
0.05	66.8491226450865\\
0.1	41.8159630835607\\
0.2	9.42794447578014\\
0.3	4.09632900118679\\
0.4	2.29658131412781\\
0.5	0.0995332064702412\\
};
  \addlegendentry{$\nu_{\text{opt}}$}
\end{axis}
\end{tikzpicture}%
%
%
\definecolor{mycolor1}{rgb}{1.00000,1.00000,0.00000}%
\begin{tikzpicture}

\begin{axis}[%
width=8.5cm,
height=2.8cm,
xmin=0,
xmax=0.5,
xlabel style={font=\color{white!15!black}},
xlabel={(b) \textit{case57}: $\text{Input domain reduction }\delta$ (--)},
ymin=-0.01,
ymax=100,
ylabel style={font=\color{white!15!black}},
ylabel={Guarantee $\nu$ (\%)},
  xtick={0, 0.1,0.2,0.3,0.4,0.5}, 
  xticklabels={0, 0.04,  0.08, 0.12, 0.16,   0.20},
axis background/.style={fill=white},
legend pos=north east,
    legend style={draw=white!15!black,legend cell align=left},
    legend columns=4
]
\addplot [color=red, mark=x, mark options={solid, red},line width =1pt]
  table[row sep=crcr]{%
0	0\\
0.01	0\\
0.05	0\\
0.1	0\\
0.2	0\\
0.3	0\\
0.4	0\\
0.5	0\\
};
\addlegendentry{$\nu_{\text{line}}$}

\addplot [color=blue, mark=x, mark options={solid, blue},line width =1pt]
  table[row sep=crcr]{%
0	100\\
0.01	94.1073886185576\\
0.05	72.4900650826934\\
0.1	48.1280370087157\\
0.2	16.5008132880681\\
0.3	2.91955662231113\\
0.4	1.60551291840251\\
0.5	0.640346704252182\\
};
\addlegendentry{$\nu_{\text{g}}$}

\addplot [color=green, mark=x, mark options={solid, green},line width =1pt]
  table[row sep=crcr]{%
0	100\\
0.01	93.5419408126352\\
0.05	69.4608425748473\\
0.1	44.0560881574027\\
0.2	12.1057376049801\\
0.3	2.34859372015915\\
0.4	0.836398670639751\\
0.5	0.333591294130036\\
};
\addlegendentry{$\nu_{\text{dist}}$}

\addplot [color=black, mark=x, mark options={solid, black},line width =1pt]
  table[row sep=crcr]{%
0	100\\
0.01	93.5399504986551\\
0.05	69.4492489389673\\
0.1	44.0349182246635\\
0.2	10.572905704866\\
0.3	1.56022445227012\\
0.4	0.200290970849035\\
0.5	0\\
};
\addlegendentry{$\nu_{\text{opt}}$}

\end{axis}

\end{tikzpicture}%
%
%
\definecolor{mycolor1}{rgb}{1.00000,1.00000,0.00000}%
\begin{tikzpicture}

\begin{axis}[%
width=8.5cm,
height=2.8cm,
xmin=0,
xmax=0.5,
xlabel style={font=\color{white!15!black}},
xlabel={(b) \textit{case118}: $\text{Input domain reduction }\delta$ (--) },
ymin=-0.01,
ymax=100,
ylabel style={font=\color{white!15!black}},
ylabel={Guarantee $\nu$ (\%)},
  xtick={0, 0.1,0.2,0.3,0.4,0.5}, 
  xticklabels={0, 0.04,  0.08, 0.12, 0.16,   0.20},
axis background/.style={fill=white},
legend pos=north east,
    legend style={draw=white!15!black,legend cell align=left},
    legend columns=4
]
\addplot [color=red, mark=x, mark options={solid, red},line width =1pt]
  table[row sep=crcr]{%
0	100\\
0.01	95.5955285257999\\
0.05	77.4142708167114\\
0.1	53.5626684232394\\
0.2	16.8871834953093\\
0.3	2.76401406098171\\
0.4	1.13976113993721\\
0.5	0.0976602342848618\\
};
\addlegendentry{$\nu_{\text{line}}$}

\addplot [color=blue, mark=x, mark options={solid, blue},line width =1pt]
  table[row sep=crcr]{%
0	100\\
0.01	95.5667861343471\\
0.05	80.3547609538637\\
0.1	61.7370539976811\\
0.2	32.4054379366467\\
0.3	16.8404344677768\\
0.4	6.93793544370729\\
0.5	0.0128761831311015\\
};
\addlegendentry{$\nu_{\text{g}}$}

\addplot [color=green, mark=x, mark options={solid, green},line width =1pt]
  table[row sep=crcr]{%
0	100\\
0.01	96.5446160441971\\
0.05	82.6838426281826\\
0.1	65.6409297086228\\
0.2	32.8052868403302\\
0.3	15.8590216637492\\
0.4	7.54557819541704\\
0.5	0.625862642822152\\
};
\addlegendentry{$\nu_{\text{dist}}$}

\addplot [color=black, mark=x, mark options={solid, black},line width =1pt]
  table[row sep=crcr]{%
0	100\\
0.01	96.7331140888872\\
0.05	83.5870615801402\\
0.1	66.927915777398\\
0.2	36.1905046879928\\
0.3	18.1620169865212\\
0.4	10.1505715256191\\
0.5	2.52748592381511\\
};
\addlegendentry{$\nu_{\text{opt}}$}

\end{axis}

\end{tikzpicture}%
    \end{footnotesize}
    \caption{The worst-case guarantees are shown as a function of the input domain reduction $\delta$ for \textit{case39}, \textit{case57} and \textit{case118}. Note that the values are normalized to 100\% with respect to the worst-case values reported in Tables~\ref{Table_WC_Constr_Viol} and \ref{Table_WC_dist} for the entire initial input domain.}
    \label{Input_domain}
     \vspace{-0.3cm}
\end{figure}
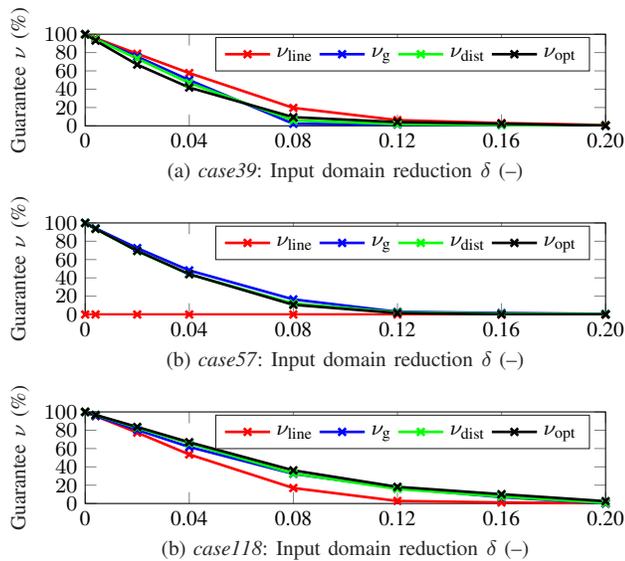
In the following, we demonstrate that the worst-case guarantees can be systematically reduced by training on a larger input domain than the worst-case guarantees are evaluated on. We achieve this by reducing the input domain $\mathcal{D}$ with a term $\delta$ that can vary between 0.0 and 0.2: $(0.6 + \delta) \bm{p_{\text{d}}^{\text{max}}} \leq \bm{p_{\text{d}}} \leq (1.0 - \delta) \bm{p_{\text{d}}^{\text{max}}}$. For \textit{case39}, \textit{case57} and \textit{case118}, Fig.~\ref{Input_domain} shows the worst-case guarantees as a function of the input domain reduction $\delta$. 
Note that the values on y-axis are normalized to 100\% with respect to the worst-case values reported in Tables~\ref{Table_WC_Constr_Viol} and \ref{Table_WC_dist} for the entire initial input domain. First, we can observe that the inputs (i.e., the loading $\mb{p_{\text{d}}}$) which lead to the worst-case performance are at the boundary of the input domain. Second, by increasing the input domain reduction $\delta$, the worst-case bounds can be systematically reduced (e.g., for these three test cases, by reducing each dimension by $\delta = 0.08$, we can reduce all worst-case guarantees to below 20\% compared to the initial domain). 
This implies that to reach an acceptable worst-case performance on a specified domain, the neural network can be re-trained on a larger domain if the initial performance is not satisfactory. 


\section{Conclusion} \label{Conclusion}
This work introduces for the first time a framework to obtain worst-case guarantees for neural networks. As a guiding example, we apply it to neural networks predicting DC-OPF solutions. Our work addresses a major barrier which, if removed, would enable the application of neural networks in safety-critical systems. Leveraging mixed-integer linear reformulations of trained neural networks, we can obtain worst-case guarantees with respect to the maximum physical constraint violations, the maximum distance between the predicted and the optimal decision variables, and the maximum sub-optimality. For a range of PGLib-OPF networks up to 300 buses, we show that the obtained worst-case guarantees can be up to one order of magnitude larger than the empirical lower bounds (i.e. computing the maximum of an error metric on the discrete samples of the entire dataset). More importantly, we show that the worst-case predictions appear on the boundaries of the input domain used for training. As a result, the worst-case guarantees can be systematically reduced by training the neural network on a larger input domain, and applying it on a subdomain. Future work is directed towards robust neural network training and obtaining worst-case guarantees for predicting solutions to AC-OPF problems. 

\bibliographystyle{IEEEtran}

\bibliography{Bib}

\end{document}